\documentclass[conference]{IEEEtran}

\usepackage{amsmath,amsfonts,amssymb}
\usepackage{graphicx}
\usepackage{url}
\usepackage{array}
\usepackage{cite}
\usepackage{hyperref}  

\begin{document}

\title{Less Context, Same Performance: A RAG Framework for Resource-Efficient LLM-Based Clinical NLP}

\author{
\IEEEauthorblockN{
Satya Narayana Cheetirala\textsuperscript{1},
Ganesh Raut\textsuperscript{1},
Dhavalkumar Patel\textsuperscript{1},
Fabio Sanatana\textsuperscript{1},
Robert Freeman\textsuperscript{1},\\
Matthew Levin\textsuperscript{2,4},
Girish N. Nadkarni\textsuperscript{3,4},
Omar Dawkins\textsuperscript{6},
Reba Miller\textsuperscript{5},\\
Randolph M. Steinhagen\textsuperscript{7},
Eyal Klang\textsuperscript{3,4},
Prem Timsina\textsuperscript{1}
}

\IEEEauthorblockA{\textsuperscript{1}Institute for Healthcare Delivery Science}
\IEEEauthorblockA{\textsuperscript{2}Department of Anesthesiology, Perioperative and Pain Medicine}
\IEEEauthorblockA{\textsuperscript{3}Hasso Plattner Institute for Digital Health}
\IEEEauthorblockA{\textsuperscript{4}Windreich Department of Artificial Intelligence and Human Health, Mount Sinai Medical Center, NY, USA}
\IEEEauthorblockA{\textsuperscript{5}Department of Surgery}
\IEEEauthorblockA{\textsuperscript{6}Department of Surgery, Senior Quality Data Analyst – ISM}
\IEEEauthorblockA{\textsuperscript{7}Department of Surgery, Division of Quality and Patient Safety, All at Icahn School of Medicine at Mount Sinai, NY, USA}

\vspace{1mm}
\IEEEauthorblockA{
Email: satyanarayana.cheetirala@mountsinai.org \\
\textbf{Code Repository:} \href{https://github.com/mountsinai/less-context-same-performance}{github.com/mountsinai/less-context-same-performance}
}
}

\maketitle

\begin{abstract}

Long text classification is challenging for Large Language Models (LLMs) due to token limits and high computational costs. This study explores whether a Retrieval-Augmented Generation (RAG) approach---using only the most relevant text segments---can match the performance of processing entire clinical notes with large-context LLMs. We begin by splitting clinical documents into smaller chunks, converting them into vector embeddings, and storing these in a FAISS index. We then retrieve the top 4,000 words most pertinent to the classification query and feed these consolidated segments into an LLM.

We evaluated three LLMs (GPT4o, LLaMA, and Mistral) on a surgical complication identification task. Metrics such as AUC ROC, precision, recall, and F1 showed no statistically significant differences between the RAG-based approach and whole-text processing \((p > 0.05p >0.05 )\). These findings indicate that RAG can significantly reduce token usage without sacrificing classification accuracy, providing a scalable and cost-effective solution for analyzing lengthy clinical documents.
\end{abstract}


\section{Introduction}
With the advent of Large Language Models (LLMs) such as GPT-4 and LLaMA, the capacity to interpret and generate human-like text has broadened the horizon of natural language processing (NLP). However, these large models still face limitations related to token length and considerable computational overhead---issues exacerbated by extended clinical documents, which can span hundreds of thousands of tokens. This problem is particularly acute in healthcare, where detailed patient records and surgical notes make large-scale LLM usage costly.

One promising avenue to mitigate these limitations is the Retrieval-Augmented Generation (RAG) framework, wherein only the most relevant chunks of text are retrieved and passed to the LLM. By restricting the model’s context to the highest-yield content (e.g., a 4{,}000-token limit), RAG can reduce computational requirements, minimize latency, and still maintain robust classification or predictive performance.

In this paper, we compare a RAG-based approach against a direct whole-text ingestion strategy for classifying post-operative complications. We hypothesize that RAG---despite using fewer tokens---can achieve equivalent performance to full-document methods.

\section{Literature Review}
\subsection{Token Constraints and Computational Costs}
Contemporary LLMs typically limit input length to a few thousand tokens, although new variants (e.g., GPT4o, Gemini) promise extended contexts of up to 128K or even 1M tokens. While appealing for clinical NLP tasks involving lengthy documents, extended context windows demand substantial computational resources and network I/O. Standard Transformer-based models scale quadratically in both time and memory with respect to input sequence length due to their self-attention mechanism~\cite{Zaheer2020bigbird,Beltagy2020longformer}. Recent variants employing optimized attention mechanisms, such as Flash Attention~\cite{Dao2022flashattention}, mitigate the quadratic memory complexity but still require substantial computational resources. Consequently, large-context language models, even with optimized architectures, remain computationally intensive and resource-demanding to train, fine-tune, deploy, and run inference, posing significant challenges for resource-constrained healthcare systems.

Moreover, merely increasing the token limit does not guarantee superior performance. Extremely long contexts can introduce noise and overwhelm attention mechanisms, potentially obscuring essential information~\cite{Sakai2025healthcareLLM}. As an alternative, selective methods like RAG provide a streamlined solution by focusing solely on clinically relevant segments, thereby reducing computational overhead while preserving accuracy~\cite{Zakka2023almanac,Unlu2024clinicalTrialScreening}.

\subsection{Retrieval-Augmented Generation (RAG)}
RAG is a framework that uses a retrieval step to fetch relevant text segments from an indexed corpus prior to feeding them into an LLM~\cite{Lewis2020rag}. This strategy narrows input to smaller, topic-specific segments, sharply cutting token usage while retaining essential information for decision-making. In clinical contexts, RAG highlights core evidence---such as details about diagnoses and post-operative complications---and has demonstrated improved efficiency and interpretability~\cite{Gupta2024compSurveyRAG}.

\subsection{Chunking vs. Whole-Text Ingestion}
\subsubsection{Chunking Approaches}
Chunk-based methods split lengthy clinical documents into smaller segments (e.g., 512 or 1{,}024 tokens) and process each individually~\cite{Qu2024semanticChunking}. The outputs may then be aggregated (e.g., via a hierarchical attention mechanism) to maintain context. While this reduces computational demands, it risks fragmenting the text. Hierarchical models attempt to address this limitation by encoding each chunk first and combining representations afterward. Nevertheless, this still requires processing the entire input text~\cite{Wang2018wordEmbeddings}.

\subsubsection{Whole-Text Ingestion}
Whole-text ingestion uses extended-context LLMs capable of handling an entire document in a single pass (e.g., GPT4o supporting 128K tokens). Research on automated ICD coding suggests that larger contexts can improve performance by incorporating more tokens~\cite{Yoo2025ICDcodingLLM}. Nevertheless, these models remain expensive to run and may be impractical for extremely long documents or high-volume clinical workloads.

\subsection{Vector Embeddings}
Modern retrieval-based pipelines typically pair dense vector embeddings with high-speed similarity search engines like FAISS(Facebook AI Similarity Search)~\cite{Johnson2019faiss}. Clinical text segments (e.g., sentences or paragraphs) are converted into embeddings using transformer-based encoders specialized for semantic similarity tasks. These embeddings are indexed in FAISS for swift approximate nearest-neighbor searches. The system then retrieves the top-\textit{N} most relevant chunks to a query, pruning irrelevant details early and potentially boosting accuracy~\cite{Beam2020clinicalConceptEmbeddings,Busolin2024earlyExitDenseRetrieval}.

\subsection{Implementing Top-N Retrieval with FAISS}
When facing lengthy documents, one can tokenize and chunk the text, embed each chunk, and index the embeddings in FAISS. A query embedding---based on a specific question or scenario---retrieves the highest-ranking chunks. These top-\textit{N} chunks are then supplied to an LLM or classifier. This retrieval-augmented workflow has shown improvements in runtime efficiency and classification performance by filtering out superfluous text~\cite{Teodoro2017improvingAvgRanking,Patel2024promptEngUSMLE}.

\section{Methodology}
\subsection{Dataset and Labeling}

This study utilized clinical patient data from the Mount Sinai Health system(MSHS), consisting of six hospitals in New York City that include a quaternary academic medical center, tertiary care urban hospitals, and community hospitals. Prior to beginning the study, we obtained a waiver of informed consent from our Institutional Review Board (IRB) (IRB-18-00573-MODCR001).

The Department of Surgery maintains a Mortality and Morbidity M/\&M data registry, which provides a comprehensive record of surgical complications. Residents initially flag potential complications, and these cases are subsequently reviewed by a dedicated quality team. The team adjudicates whether each flagged event constitutes a true surgical complication. In this study, we retrieved confirmed complications directly from this M/\&M database. To form the negative (i.e., no-complication) cohort, we included patients who underwent surgery within the same health system but had no entries in the M/\&M registry. A patient was considered a negative case if there was no evidence of complication documented in the M/\&M surgical complication database.

Figure 1 depicts the dataset selection approach. We selected our study sample from records covering 2013 through 2023. Specifically, we randomly picked 996 positive cases out of a total 4,248 available positive cases from M/\&M database, and 1,298 negative cases from a pool of 344,505. Altogether, our final group comprised 2,294 individuals who had surgical procedures in the healthcare system.

\begin{figure}[ht]
\centering
\includegraphics[width=2.9in]{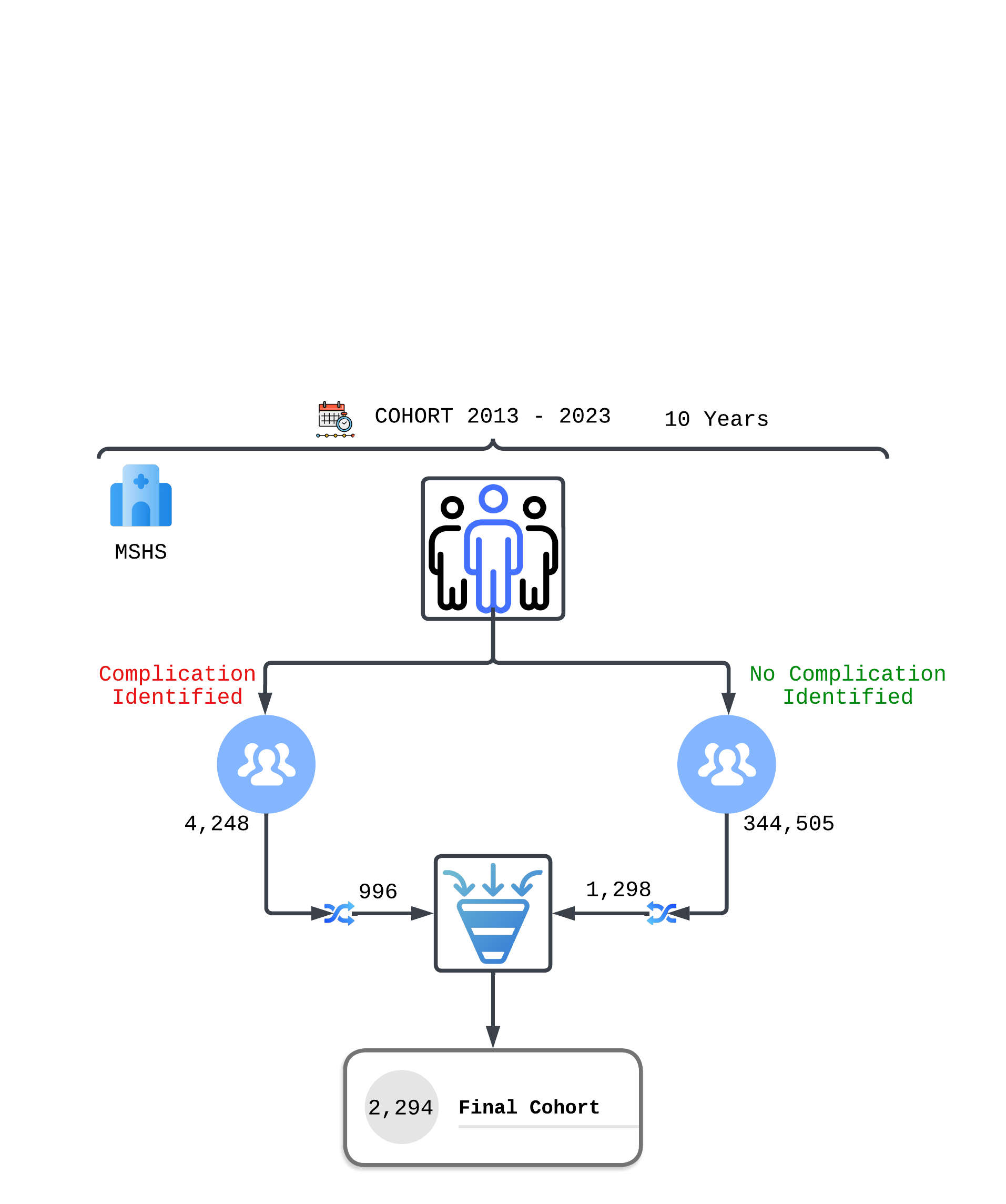}  
\caption{Overview of the study cohort. We randomly selected 996 cases with surgical complications (2013--2023) and 1,298 cases with no identified complications.}
\label{fig:study_cohort}
\end{figure}

\subsection{Study Design}
This study aimed to identify surgical complications within the MSHS by classifying them as either positive or negative. Two approaches were explored:  Two approaches were compared:

\begin{itemize}
    \item \textbf{A1-Long Context (Whole-Text Ingestion):} An LLM processes entire clinical notes in one pass.
    \item \textbf{A2-RAG (Retrieval-Augmented Generation):} An LLM processes only the top retrieved text segments, concatenated into a shorter context.
\end{itemize}

\begin{figure}[ht]
\centering
\includegraphics[width=3.0in]{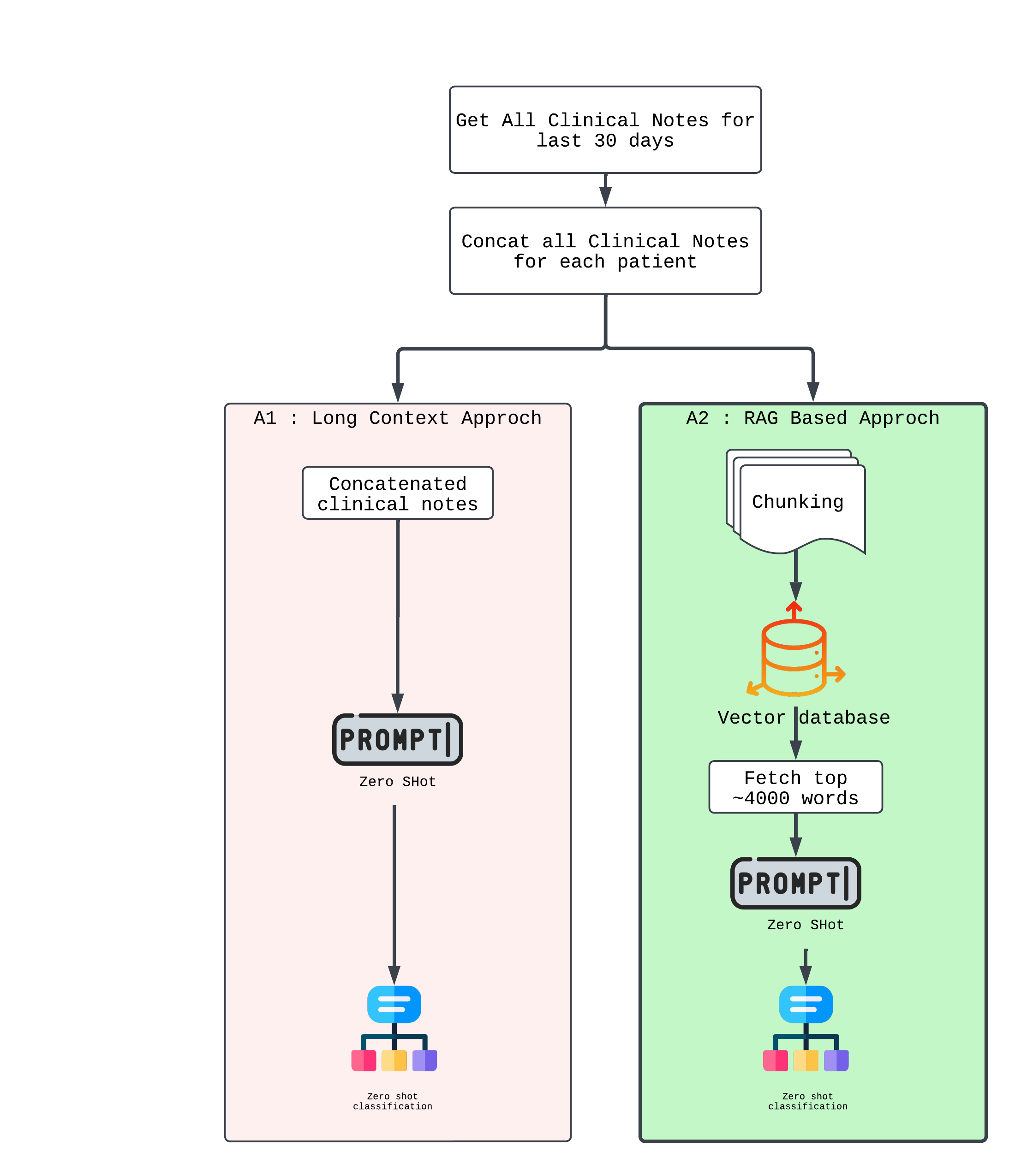} 
\caption{Overall study design comparing a long-context approach vs. a RAG approach for classifying post-operative complications.}
\label{fig:study_design}
\end{figure}

\subsubsection{A1-Long Context Approach}
\begin{figure}[ht]
\centering
\includegraphics[width=3.2in]{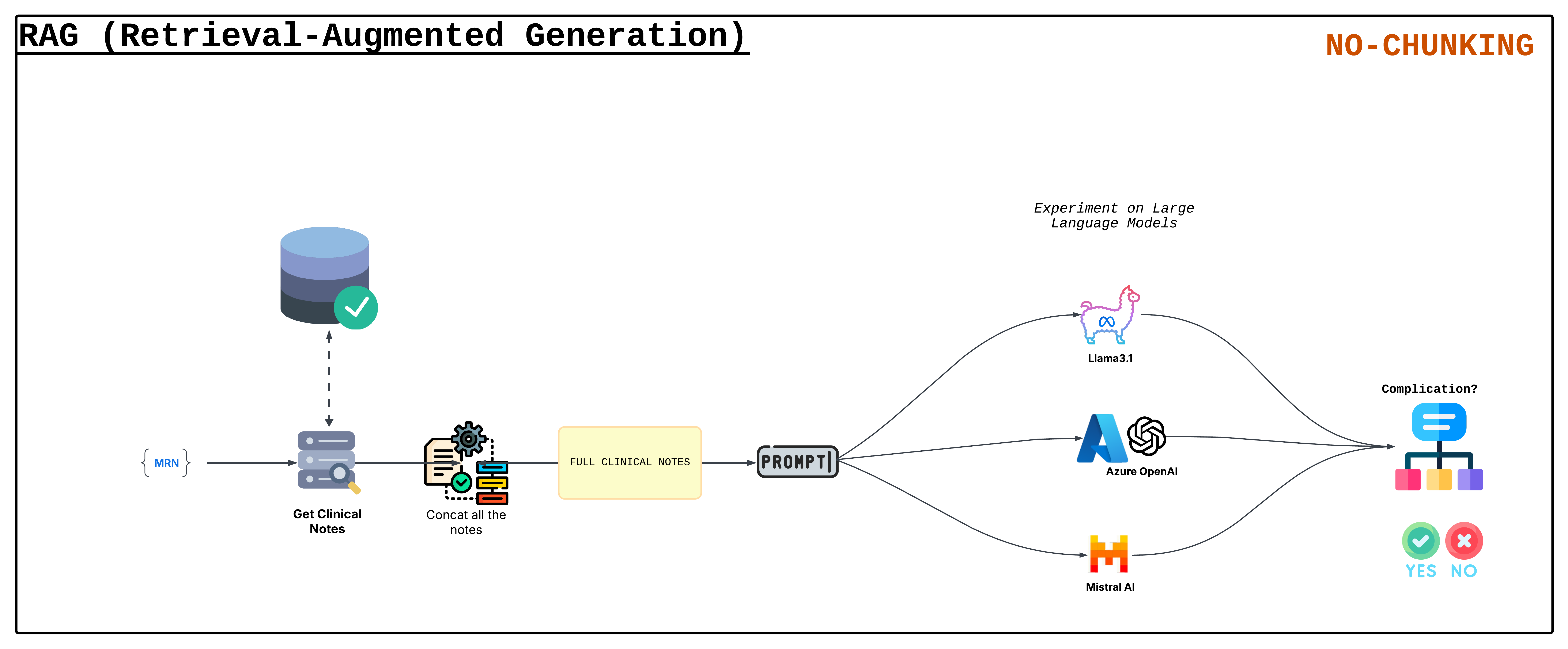} 
\caption{Long Context Approach. Entire text from the past 30 days of clinical notes is concatenated and fed to the LLM in a single pass.}
\label{fig:long_context}
\end{figure}

As depicted in Figure~\ref{fig:long_context}, we query the system to retrieve the last 30 days of clinical notes for each patient using their Medical Record Number (MRN). All relevant notes are concatenated into a single text block, which is then fed into the LLM. Three models were tested under this whole-text configuration (see the GitHub code repository).

\subsubsection{A2-RAG Approach}
\begin{figure}[ht]
\centering
\includegraphics[width=3.2in]{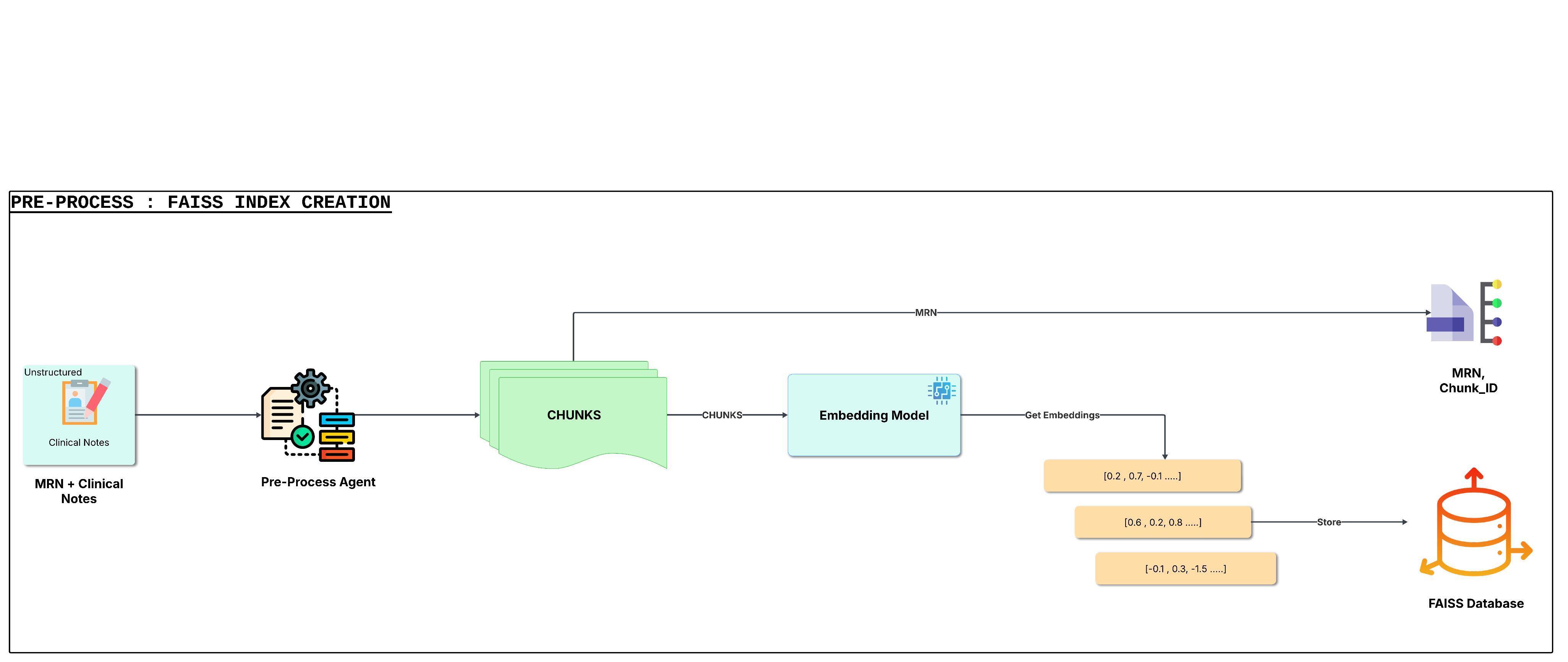} 
\caption{Vector database creation using FAISS. Clinical notes are chunked, embedded, and indexed for efficient similarity searches.}
\label{fig:db_creation}
\end{figure}

Our second approach is divided into two major steps: 1) building a vector database using FAISS (Figure~\ref{fig:db_creation}) to efficiently manage large-scale text data and 2) applying a Retrieval-Augmented Generation (RAG) framework to retrieve relevant information from this database and classify post-operative complications using a shortened context.

\paragraph{Vector Database Creation  }
In this initial phase, each clinical note is split into chunks of up to 512 words, ensuring manageable segment sizes for embedding generation. These chunks are then transformed into vector embeddings using a transformer-based encoder. To enable efficient similarity searches, the resulting embeddings are stored in a FAISS index. This approach supports rapid retrieval in high-dimensional spaces and simplifies large-scale data handling for downstream tasks.

\begin{figure}[ht]
\centering
\includegraphics[width=3.8in]{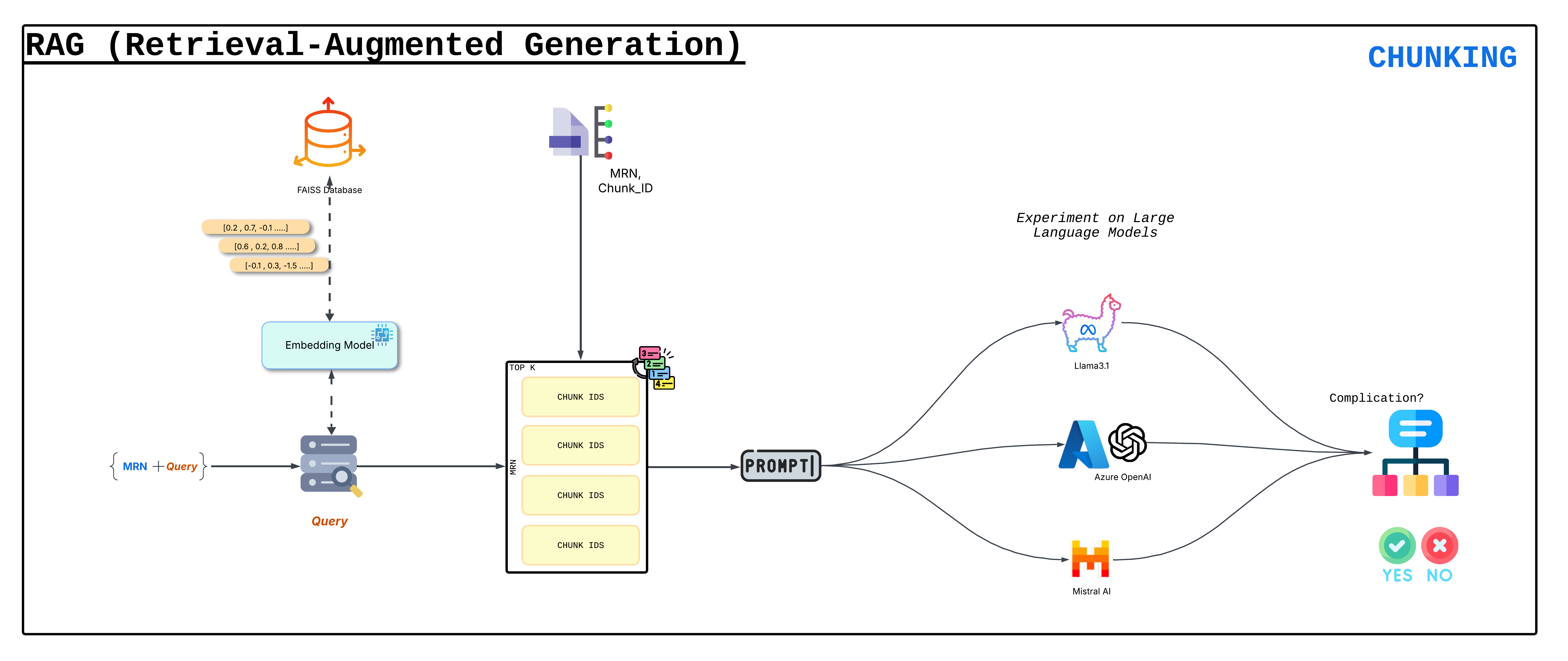} 
\caption{RAG-based classification. The top-$N$ chunks from the FAISS index are concatenated (under 4,000 tokens) and fed into an LLM classifier.}
\label{fig:rag_classification}
\end{figure}

\paragraph{Classification with RAG (Figure~\ref{fig:rag_classification})}
Once the vector database is established, a targeted prompt is used to identify segments potentially indicating post-operative complications[16]. The FAISS index retrieves the top N most semantically similar chunks, which are then concatenated in their original order to preserve narrative flow. The combined text is kept under a 4,000-token limit to respect the model’s context window. Finally, this text block is fed into the LLM-based classifier, which assesses the presence or absence of complications, assigns a severity score, and generates a binary label (e.g., 0 or 1) to quantify clinical significance.

\subsection{Experimental Setup}
To compare the RAG-based short context with the whole-text long context for surgical complication classification, we evaluated three distinct Large Language Models (LLMs):
\begin{itemize}
    \item \textbf{GPT4o:} A proprietary model with an extended context window, facilitating the processing of more tokens per prompt.
    \item \textbf{LLaMA (llama3.1:8b-instruct-fp16):} A smaller, open-source model noted for its accessibility and ease of fine-tuning.
    \item \textbf{Mistral (mistral-nemo:12b-instruct-2407-q2\_K):}Another open-source model optimized for computational efficiency while maintaining competitive performance.
\end{itemize}

Each model was deployed under two primary modes:
1.	A1-Long Context Approach (Whole-Text Ingestion): An LLM processes entire clinical notes in one pass.
2.	A2-RAG Approach (Retrieval-Augmented Generation): An LLM processes only the top retrieved text segments concatenated into a shorter context.
All three models were tested on the same set of patient records to ensure comparability. For each mode, the model was tasked with classifying the presence or absence of post-operative complications, assigning a severity score, and producing a binary label (0 or 1). Relevant hyperparameters, including retrieval chunk size and the maximum number of chunks, were held constant for consistency.

\subsection{Infrastructure Setup}

All experiments were conducted on a HIPAA-compliant Azure infrastructure to ensure data security and regulatory compliance. For models provided by OpenAI, we utilized HIPAA-compliant Azure OpenAI endpoints. For open-source models such as LLAMA and Mistral, we employed the OLAMA framework, deploying both on secure Azure-hosted virtual machines.
The virtual machine used was a “Standard NC40ads H100 v5” instance, configured with 40 vCPUs and 320 GiB of RAM—well-suited for high-performance computing and GPU-intensive workloads. This VM runs a 64-bit Linux operating system (V2 generation) and includes a SCSI-based disk controller, an encrypted OS disk, and one attached data disk to support efficient data text processing~\cite{patel2024cloud}.

\section{Results}

\begin{figure*}[t]
    \centering
    \includegraphics[width=0.95\textwidth]{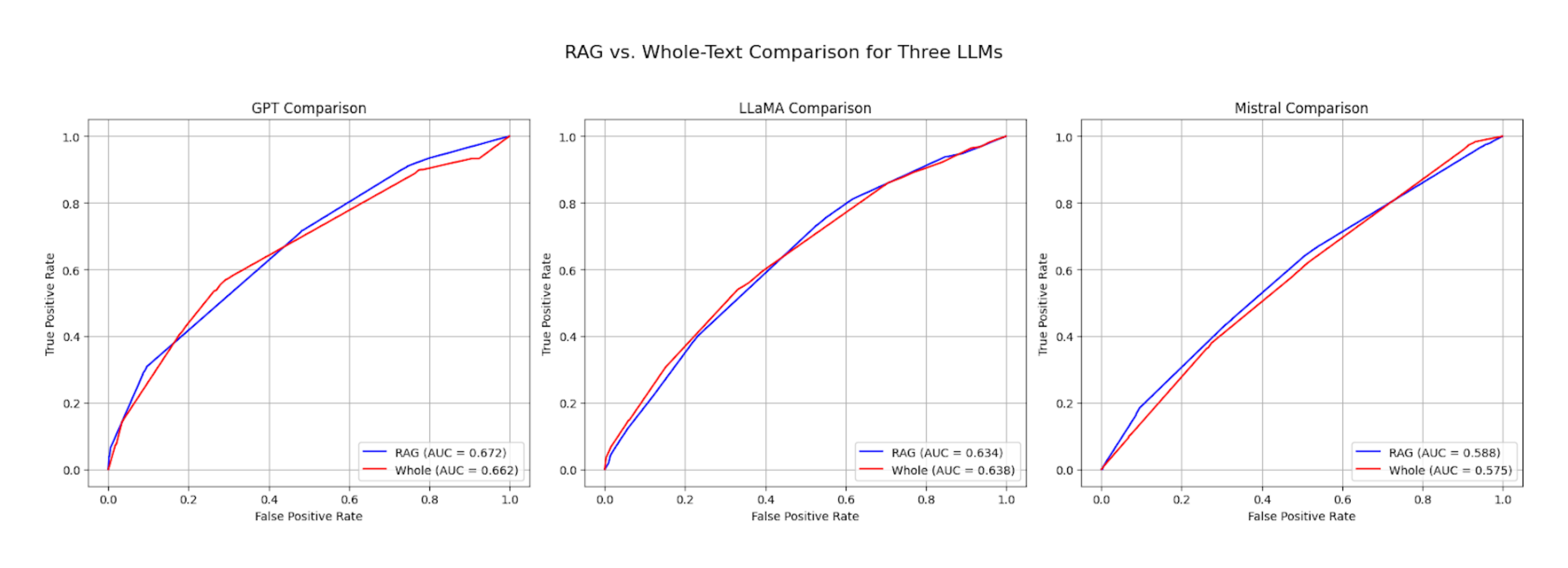}
    \caption{RAG vs. Whole-Text Comparison for Three LLMs. The blue curves represent the RAG approach and the red curves represent the Whole-Text approach.}
    \label{fig:rag_whole_comparison}
\end{figure*}

As shown in Fig.~\ref{fig:rag_whole_comparison}, the AUROC curves demonstrate that the RAG-based ingestion method performs comparably to whole-text ingestion across all three evaluated LLMs. The blue curves correspond to the RAG approach, while the red curves represent whole-text ingestion. Across models, the curves align closely, indicating minimal loss in discriminative performance when using a reduced context.

Quantitative results are summarized in Table~\ref{tab:performance}. For each model, we report AUROC, precision, recall, F1 score, and PR AUC. The GPT4o-RAG and GPT4o-Long Context approaches yielded nearly identical F1 scores (0.61), despite the RAG method using significantly fewer tokens. Similar patterns were observed for the LLaMA and Mistral models.

\begin{table}[ht]
\centering
\caption{Performance Results for Post-Operative Complication Classification}
\label{tab:performance}
\resizebox{\columnwidth}{!}{%
\begin{tabular}{lccccc}
\hline
\textbf{Experiment} & \textbf{AUROC} & \textbf{Precision} & \textbf{Recall} & \textbf{F1} & \textbf{PR AUC} \\
\hline
GPT4o-RAG            & 0.67        & 0.53              & 0.71           & 0.61       & 0.64 \\
GPT4o-Long Context   & 0.66        & 0.46              & 0.90           & 0.61       & 0.61 \\
LLaMA-RAG            & 0.63        & 0.51              & 0.73           & 0.60       & 0.55 \\
LLaMA-Long Context   & 0.63        & 0.48              & 0.86           & 0.61       & 0.58 \\
Mistral-RAG          & 0.58        & 0.44              & 0.96           & 0.60       & 0.51 \\
Mistral-Long Context & 0.57        & 0.44              & 0.96           & 0.61       & 0.49 \\
\hline
\end{tabular}%
}
\end{table}

\subsection{Cost and Resource Analysis}

\begin{figure}[ht]
    \centering
    \includegraphics[width=3.2in]{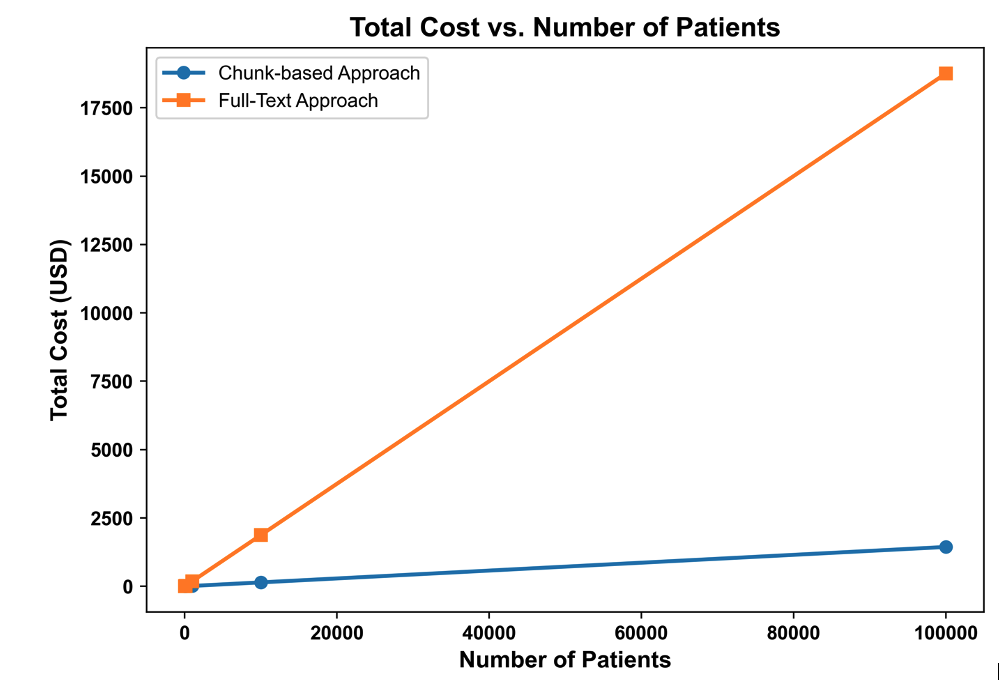}
    \caption{Total Cost vs. Number of Patients.}
    \label{fig:totalcost}
\end{figure}

To evaluate efficiency, we calculated token usage and associated costs. For all 2,293 patients, the full-text approach with GPT-4o generated approximately 172 million tokens, translating to \$430 in API usage (at \$2.50 per million tokens). In contrast, the chunk-based RAG method required only 13.2 million tokens, costing approximately \$33—representing a cost reduction of over 90\%. Fig.~\ref{fig:totalcost} shows the projection of cost for 100K patients.

\begin{figure}[ht]
    \centering
    \includegraphics[width=3.2in]{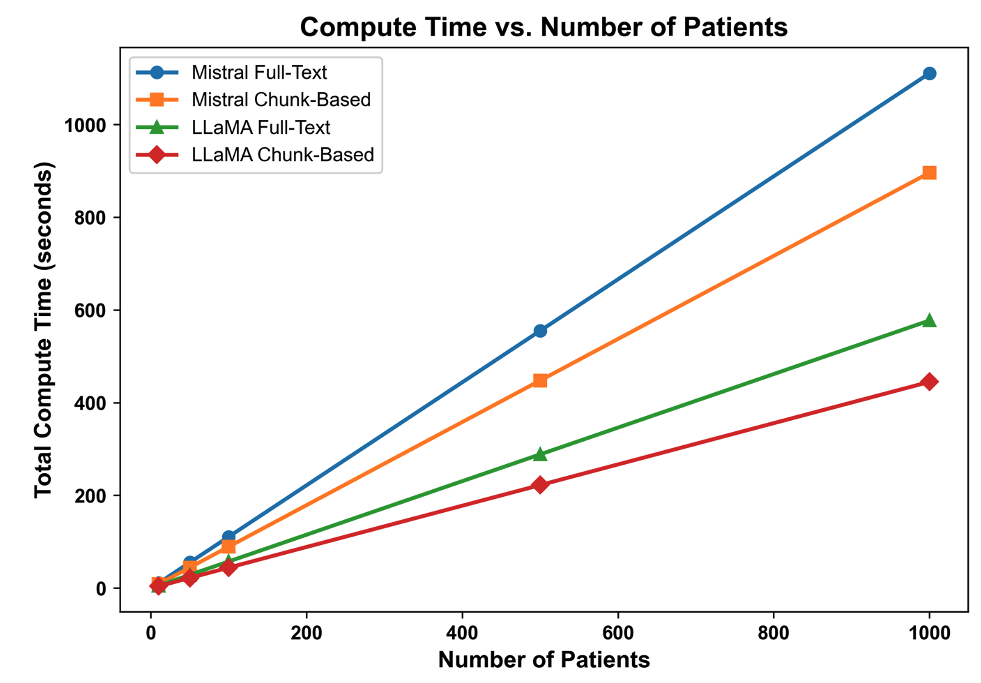}
    \caption{Compute Time vs. Number of Patients.}
    \label{fig:compute_time}
\end{figure}

Time benchmarks for 100 patients were evaluated for open-source models(LLAMA and Mistral). Mistral achieved a 19\% speed improvement in chunk-based mode (from 1.11 to 0.90 seconds per patient), while LLaMA improved by 23\% (from 0.58 to 0.45 seconds per patient). This affirms that the chunking process not only saves cost but also improves inference speed. Fig.~\ref{fig:compute_time} shows the projection of runtime for 1000 patients.

\subsection{DeLong Test: RAG vs. Long Context}

To determine whether AUROC differences between the RAG and full-text methods were statistically significant, we performed DeLong tests. As shown in Table~\ref{tab:delong}, all $p$-values did not exceeded the 0.05 threshold, indicating that the observed performance differences were not statistically significant.

\begin{table}[ht]
\centering
\caption{DeLong Test Results Comparing RAG vs. Whole-Text Ingestion}
\label{tab:delong}
\begin{tabular}{lc}
\hline
\textbf{Model} & \textbf{p-value} \\
\hline
GPT4o          & 0.33 \\
LLaMA          & 0.78 \\
Mistral        & 0.38 \\
\hline
\end{tabular}
\end{table}

\section{Discussion}

\subsection{ Summary of Key Findings}
This study demonstrates that a Retrieval-Augmented Generation (RAG) approach can match the classification performance of a long-context LLM that ingests entire clinical notes. Notably, this equivalence holds across three distinct models—GPT4o, LLaMA, and Mistral—underscoring RAG’s adaptability in varying computational environments. Moreover, the results remain consistent even when comparing models of different complexities and sizes, highlighting RAG’s scalability and potential for diverse clinical applications. DeLong test results further confirm that differences in AUC ROC between the two ingestion strategies were statistically insignificant.

\subsection{ Implications for Clinical NLP}

\subsubsection{ Token Efficiency and Computational Savings}
By focusing on high-yield text segments, RAG conserves computational resources without degrading classification accuracy. Transformer-based models typically incur substantial costs as input length grows; thus, RAG’s token-efficiency is particularly advantageous in healthcare, where large-scale or real-time applications must operate under strict computational and cost constraints.

\subsubsection{ Preservation of Classification Accuracy}
Despite discarding a large portion of clinical notes, the RAG-based pipeline maintained statistical parity with whole-text ingestion. This suggests that much of the text in clinical notes may be redundant for certain classification tasks (e.g., identifying post-operative complications) and that targeted retrieval of key segments can suffice. For healthcare decision-support systems, RAG provides a viable path to scaling NLP solutions without requiring extremely large context windows.

\subsubsection{ Potential for Broader Applicability}
While this study focused on surgical complication identification, the RAG approach can extend to other tasks in clinical NLP—such as ICD coding, radiology report classification, or medication extraction—where domain-specific signals are concentrated in particular text regions. Moreover, the method is model-agnostic; any transformer-based architecture that benefits from chunked inputs and vector retrieval can integrate this strategy.

\subsection{ Practical Considerations}

\subsubsection{ Indexing Infrastructure}
Although indexing incurs initial overhead (for creating and maintaining FAISS indexes), it confers significant downstream benefits. Systems can adapt the same index for multiple tasks, adjusting only the retrieval query or prompt without re-embedding the entire corpus.

\subsubsection{ Model and Context Window Limitations}
Large-context LLMs (e.g., GPT4o, Gemini) remain costly, both in licensing and hardware. RAG-based pipelines constrain context to a manageable size (e.g., 4{,}000 tokens) and thus avoid the steep scaling costs associated with extremely long sequence lengths.

\subsubsection{ Interoperability and Quality Assurance}
RAG enhances interoperability by pinpointing the precise chunks responsible for classification, potentially improving clinician trust. Nevertheless, human oversight remains necessary, particularly in high-stakes decisions like patient care and risk assessments.

\section{Conclusion}
This study confirms that a Retrieval-Augmented Generation approach, which filters large clinical notes to their most relevant segments, can match the performance of processing an entire note at once. Given the computational and financial constraints of large context windows, RAG stands out as an efficient alternative with no observed drop in accuracy or recall.

\noindent\textbf{Future Work}
\begin{enumerate}
    \item \textbf{Sophisticated Retrieval:} Investigate advanced retrieval techniques (e.g., hierarchical chunking, query expansion) for scenarios demanding additional contextual depth.
    \item \textbf{Application Expansion:} Extend this RAG framework to tasks such as automated ICD coding, radiology report classification, and real-time patient triage.
    \item \textbf{Cost-Benefit Analyses:} Perform detailed economic evaluations to guide resource-limited healthcare systems in deciding between indexing overhead and on-demand inference costs.
\end{enumerate}

\bibliographystyle{IEEEtran}

\appendices
\renewcommand{\arraystretch}{1.1} 

\section{Note Types}
\label{appendixA}
\begin{table}[ht]
\centering
\footnotesize
\setlength{\tabcolsep}{5pt}
\caption{Note Types}
\begin{tabular}{ll}
\hline
\textbf{SN} & \textbf{Note Type} \\
\hline
1  & IP Operative Report \\
2  & Addendum IP Operative Report \\
3  & OP Operative Report \\
4  & Addendum OP Operative Report \\
5  & Brief Op Note \\
6  & Perioperative Record \\
7  & OR PreOp \\
8  & OR PreOp Anesthesia \\
9  & Pre-Op Medical Assessment \\
10 & OR PostOp \\
11 & Anesthesia Procedure Notes \\
12 & Anesthesia Preprocedural Evaluation \\
13 & Anesthesia Postprocedural Evaluation \\
14 & OR Nursing \\
15 & Anesthesia Transfer of Care \\
16 & Anesthesia PACU Discharge \\
\hline
\end{tabular}
\end{table}
\vspace{-2mm}

\section{Code Repo}
\label{appendixB}
\noindent
The complete code for our RAG pipeline and experiments is available at:
\href{https://github.com/mountsinai/less-context-same-performance}{\url{https://github.com/mountsinai/less-context-same-performance}}

\vspace{-2mm}
\section{Glossary}
\label{appendixC}

Below are brief definitions of key terms used in this study:

\begin{table}[ht]
\centering
\footnotesize
\setlength{\tabcolsep}{4pt}
\caption{Core Concepts and Frameworks}
\begin{tabular}{p{1.1cm} p{2.8cm} p{3.5cm}}
\hline
\textbf{Term} & \textbf{Full Name} & \textbf{Definition} \\
\hline
RAG & Retrieval-Augmented Generation & 
A framework that uses a retrieval step to fetch relevant text segments from an indexed corpus prior to feeding them into an LLM. \\
NLP & Natural Language Processing & 
The field of computer science and AI concerning interactions between computers and human language. \\
LLM & Large Language Model & 
Advanced AI models (e.g., GPT-4) that generate human-like text but have token-limit constraints. \\
Token & Token & 
The basic unit of text that language models process. \\
Context Window & Context Window & 
The max text length (in tokens) an LLM can process at once. \\
\hline
\end{tabular}
\end{table}
\vspace{-2mm}

\begin{table}[ht]
\centering
\footnotesize
\setlength{\tabcolsep}{3.5pt}
\caption{Technical Terms and Methodologies}
\begin{tabular}{p{1.6cm} p{3.3cm} p{3.3cm}}
\hline
\textbf{Term} & \textbf{Full Name} & \textbf{Definition} \\
\hline
FAISS & Facebook AI Similarity Search & 
A library enabling efficient similarity searches in high-dimensional spaces. \\
Vector Embeddings & Vector Embeddings & 
Numerical representations of text capturing semantic meaning. \\
Chunking & Chunking & 
Splitting lengthy documents into smaller segments (e.g., 512 tokens). \\
Transformer-based Encoders & Transformer-based Encoders & 
Models that convert text segments into vector embeddings. \\
Whole-Text Ingestion & Whole-Text Ingestion & 
Processing an entire document in one pass. \\
Self-attention Mechanism & Self-attention Mechanism & 
A Transformer component that weighs importance of tokens. \\
FlashAttention & FlashAttention & 
An optimized attention mechanism that reduces memory overhead. \\
Hierarchical Attention Mechanism & Hierarchical Attention Mechanism & 
Encodes each chunk and merges them afterward for large documents. \\
\hline
\end{tabular}
\end{table}
\vspace{-2mm}

\begin{table}[ht]
\centering
\footnotesize
\setlength{\tabcolsep}{3.5pt}
\caption{Clinical and Healthcare Terms}
\begin{tabular}{p{0.8cm} p{3.7cm} p{3.6cm}}
\hline
\textbf{Term} & \textbf{Full Name} & \textbf{Definition} \\
\hline
M\&M & Mortality and Morbidity & 
Data registry for surgical complications. \\
MRN & Medical Record Number & 
A unique patient identifier in a healthcare system. \\
ICD Coding & International Classification of Diseases Coding & 
System for classifying diagnoses and procedures. \\
Surgical Complication & Surgical Complication & 
An adverse outcome following a surgical procedure. \\
\hline
\end{tabular}
\end{table}
\vspace{-2mm}

\begin{table}[ht]
\centering
\footnotesize
\setlength{\tabcolsep}{3.5pt}
\caption{Models Used in the Study}
\begin{tabular}{p{1.2cm} p{3.3cm} p{3.4cm}}
\hline
\textbf{Term} & \textbf{Full Name} & \textbf{Definition} \\
\hline
GPT4o & GPT4o & 
A proprietary model with an extended context window for more tokens. \\
LLaMA & llama3.1:8b-instruct-fp16 & 
A smaller, open-source model known for accessibility. \\
Mistral & mistral-nemo:12b-instruct-2407-q2\_K & 
Open-source model optimized for efficiency. \\
\hline
\end{tabular}
\end{table}
\vspace{-2mm}

\begin{table}[ht]
\centering
\footnotesize
\setlength{\tabcolsep}{2.5pt}
\caption{Study Design Approach}
\begin{tabular}{p{2.7cm} p{2.7cm} p{2.8cm}}
\hline
\textbf{Term} & \textbf{Full Name} & \textbf{Definition} \\
\hline
A1-Long Context Approach & Whole-Text Ingestion &
LLM processes entire clinical notes in one pass. \\
A2-RAG Approach & Retrieval-Augmented Generation &
LLM processes only the top retrieved text segments. \\
\hline
\end{tabular}
\end{table}

\end{document}